\documentclass[11pt]{article}
\usepackage[utf8]{inputenc}
\usepackage[english]{babel}
\usepackage{fontenc}
\usepackage{lscape}
\usepackage{mathrsfs} 
\usepackage{amsmath}
\usepackage{fullpage}
\usepackage[hidelinks]{hyperref}
\usepackage{tabularx}
\usepackage{mathtools}
\usepackage{multirow}
\usepackage{graphicx}
\usepackage{wrapfig}
\usepackage{longtable}
\usepackage{algorithm} 
\usepackage{algorithmicx}
\usepackage{algpseudocode}
\usepackage{float}

\title{Solving the clustered traveling salesman problem with $d$-relaxed priority rule} 
\begin{document}
\begin{center}

\begin{LARGE}
Solving the clustered traveling salesman problem with $d$-relaxed priority rule
\end{LARGE}

\vspace*{01cm}

\textbf{Hoa Nguyen Phuong, Huyen Tran Ngoc Nhat, Minh Ho\`ang H\`a} \\
ORLab, University of Technology and Engineering, Vietnam National University \\
\vspace*{0.2cm}
\textbf{Andr\'e Langevin, Martin Tr\'epanier}\\
Interuniversitary Research Centre on Entreprise Networks, Logistics and Transportation (CIRRELT), Polytechnique Montr\'eal\\

\vspace*{0.4cm}

\end{center}

\noindent
\textbf{Abstract.} The Clustered Traveling Salesman Problem with a Prespecified Order on the Clusters, a variant of the well-known traveling salesman problem is studied in literature. In this problem, delivery locations are divided into clusters with different urgency levels and more urgent locations must be visited before less urgent ones. However, this could lead to an inefficient route in terms of traveling cost. This priority-oriented constraint can be relaxed by a rule called $d$-relaxed priority that provides a trade-off between transportation cost and emergency level. Our research proposes two approaches to solve the problem with $d$-relaxed priority rule. We improve the mathematical formulation proposed in the literature to construct an exact solution method. A meta-heuristic method based on the framework of Iterated Local Search with problem-tailored operators is also introduced to find approximate solutions. Experimental results show the effectiveness of our methods.
\vspace*{0.2cm}



\section{Introduction}\label{sec:intro}

The Traveling Salesman Problem (TSP) is one of the most studied optimization problems. Its objective is to find an optimal vehicle route in order to visit a given set of locations. In the TSP, the locations are supposed to have the same degrees of urgencies, i.e., they can be visited in any order. However, in a number of real-world routing applications, different levels of priorities at the delivery locations need to be taken into account in routing plans. For example, as a result of a natural disaster such as a storm, earthquake, tsunami, or hurricane, there are demands at many locations for relief supplies such as food, bottled water, blankets, or medical packs. Some locations are in more urgent need of supplies than other locations due to the relative position of the source of disasters, the damage status, or its importance (schools, hospitals, and government institutions should be considered as more important). Locations requiring the same level of urgency can be clusterized into groups. And the priority of a group during the relief process has to be considered, e.g., higher priority groups should be visited before others.

In the example above, the priorities indicate the importance (or urgency) of the demand at each location. Typically, priority 1 nodes must be served before priority 2 nodes, priority 2 nodes must be served before priority 3 nodes, and so on. Such a problem is called the Clustered Traveling Salesman Problem with a Prespecified Order on the Clusters (CTSP-PO) and has been studied in \cite{potvin1998, ahmed}. However, this rule is strict with respect to the priority and can lead to an inefficient route in terms of traveling cost.  It may be relevant to visit some lower priority nodes while serving higher priority nodes.  In \cite{pancham-thesis, pancham-paper}, the authors proposed a simple, but elegant rule called $d$-relaxed priority that provides flexibility to the decision maker in terms of capturing trade-offs between total distance and node priorities.

In \cite{pancham-paper} and Chapter 14 of \cite{vrpbook}, the $d$-relaxed priority rule is defined as follows. Given a positive number $d$, at any point of the route, if $p$ is the highest priority class among all unvisited locations, the relaxed rule allows the vehicle to visit locations with priority $p, p+1, ..., p+d$ before visiting all locations in class $p$. By changing the value of $d$, we can flexibly control to focus more on economic aspect or urgency level. Indeed, if we consider the 0-relaxed priority rule (i.e., $d = 0$), all the higher priority nodes must be visited before lower priority nodes. The problem is a CTSP--PO, the strictest version w.r.t priority. On the other hand, if $d$ is set to $g - 1$, where $g$ is the number of priorities, the problem becomes a typical TSP, all the node priorities being ignored. 

In the following, we denote the considered problem as Clustered Traveling Salesman Problem with $d$-relaxed priority rule (CTSP-$d$ for short). More formally, the problem is defined as follows. We are given an undirected graph $G = (V,E)$ in which $V$ is a set of vertices and each edge \textit{e} $\in$ \textit{E} with two endpoints defined on $V$ has associated traveling cost $c_{e}$ and traveling time $t_e$. The set $V$ of $n$ vertices \{1, 2,..., $n$\} is divided into $g$ disjoint groups; each group $V_p$ is associated to a priority $p$. Vertex 1 is a depot. The objective of the CTSP-$d$ is to find a Hamiltonian tour with minimum cost starting from and ending at the depot, such that each vertex is visited exactly once and the $d$-relaxed priority rule is satisfied.

The CTSP-$d$ has natural applications in humanitarian relief routing (see \cite{pancham-paper,pancham-thesis}). Other potential applications of the problem mentioned in \cite{pancham-paper} involve the routing of service technicians and the unmanned aerial vehicle (UAV) routing problem in which target priorities are important. In addition, the problem has applications in the distribution of commercial products in which the priority of a delivery location is determined by the storage level. Out-of-stock locations should be considered to be more important than the others and clustered into the group with the highest priority.

As mentioned above, the closest problem to the considered problem is the CTSP--PO. In fact, the CTSP--PO is equivalent to a CTSP-$d$ if the value of $d$ is set to 0. It was first introduced and solved by a genetic algorithm in \cite{potvin1998}. In 2014, Ahmed \cite{ahmed} developed a hybrid genetic algorithm using sequential constructive crossover (SCX), 2-opt search, and a local search for obtaining heuristic solution to the problem. 

A more relaxed version of the CTSP--PO is the Clustered TSP (CTSP). Here, delivery locations within the same cluster must be visited consecutively but there is no priority associated to a cluster. This leads to the fact that we can visit clusters in any order. The CTSP has been widely studied in the literature. The problem was firstly studied in \cite{chisman}. A branch-and-bound approach was proposed to solve the problem exactly. The author also showed that the CTSP can be transformed into a TSP by adding an arbitrarily large constant $M$ to the cost of each intercluster edge. Later on, a number of approaches have been introduced for solving the problem: genetic algorithm \cite{potvin1996}, tabu search \cite{laporte}, GRASP \cite{mestria2013}, and a combination of local search, GRASP, and variable neighborhood random descent \cite{mestria2018}. 

More recently, \cite{zhang} introduced a new variant of CTSP, called tabu clustered TSP (TCTSP). Nodes in TCTSP are now partitioned into two kinds of subsets: clusters and tabu node sets; exactly one node in each tabu node set has to be visited and the nodes within a same cluster are visited consecutively. The TCTSP can model a class of telemetry tracking and command (TTC) resources scheduling problem. Its goal is to efficiently schedule the TTC resources in order to enable the satellites to be operated normally in their designed orbits. Two metaheuristics combined with path relinking were proposed to solve the problem: Ant Colony Optimization (ACO) and GRASP. 

Despite its potential importance, the number of studies on the CTSP-$d$ is quite scarce. To the best of our knowledge, the first works on the problem are \cite{pancham-thesis, pancham-paper}. The CTSP-$d$ was also mentioned in Chapter 14 of \cite{vrpbook}. In \cite{pancham-paper}, the authors introduced the problem and its applications. They also derived worst-case bounds for the CTSP-$d$ with respect to the classical TSP and were able to show that the bounds are tight. The problem and its several extensions were formulated as mixed integer programs in \cite{pancham-thesis} where HTSP instances with only 30 or so nodes were solved to optimality using CPLEX. The lack of efficient solution methods and further analysis on the $d$-relaxed priority rule on larger instances brings strong motivation for this research.

The contributions of this research are as follows. First, we improve the Mixed Integer Programming (MIP) model introduced in \cite{pancham-thesis} to propose an exact method that can solve small and medium instances. We tested it on instances up to 50 nodes. Secondly, we propose the first metaheuristic to efficiently deal with large instances up to 200 nodes. Our metaheuristic combines the ideas of Greedy Randomized Adaptive Search Procedure (GRASP), Iterated Local Search (ILS) and Variable Neighborhood Descent (VND) proposed in \cite{silva} with problem-tailored operators. More precisely, we use additional structure data that allow to check the feasibility of local search moves w.r.t the $d$-relaxed priority rule in $O(1)$. And finally, computational results on different types of instances give insights on the performance of the methods, as well as the trade-off between the traveling cost and the priority.

The remainder of the paper is organized as follows. We discuss Mixed Integer Programing (MIP) formulations for the problem in Section 2. A metaheuristic is developed in Section 3. Computational experiments are presented in Section 4. And finally, Section 5 concludes our work.

\section{Formulations}\label{sub:mip}
In this section, we formulate the CTSP-$d$ as a Mixed Integer Programming (MIP). As proposed in the formulation of \cite{pancham-thesis}, we use two types of variables: (i) real variables $s_i$ representing the arrival time of the vehicle at node $i$; (ii) binary variables $x_{ij}$ equal to 1 if vehicle travels from $i$ to $j$, to 0 otherwise. Then, the CTSP-$d$ can be formulated as follows:

\begin{align}
\textrm{Minimize} \, \qquad \sum_{i, j \in V}&c_{ij}x_{ij} \label{eq:obj}\\
\textrm{subject to} \qquad \;\; \sum_{i \in V} x_{ij} &= 1 \qquad \forall j \in V \label{eq:c1}\\
\sum_{j \in V} x_{ij} &= 1 \qquad \forall i \in V \label{eq:c2}\\
s_1 &= 0 \label{eq:c3}\\
s_i + t_{ij} - M(1-x_{ij}) &\leq s_j \qquad \forall \{i,j\} \in E; j \neq 1 \label{eq:c4}\\
s_i &< s_j \qquad \forall i \in V_p, j \in V_q; p, q \in \{1,...,g\}: q > p + d \label{eq:c5}\\
x_{ij} &\in\{0,1\}  \qquad \forall i,j \in V \label{eq:c6}\\
s_i &\in R  \qquad \forall i \in V \label{eq:c7}
\end{align}

In the following, we denote this formulation as (F1). The objective function (\ref{eq:obj}) is to minimize the total traveling cost. Constraints (\ref{eq:c1}) and (\ref{eq:c2}) require that the vehicle must enter and leave each node exactly once. Constraint (\ref{eq:c3}) initializes the service at the depot. Constraints (\ref{eq:c4}) represent the relationship between variables $s$ and $x$. More precisely, they compute the starting service time at a node $j$ based on node $i$ if the vehicle travels from $i$ to $j$. Here, $M$ is a very large number. Next, $d$-relaxed constraints are modeled by (\ref{eq:c5}) and will be explained in the following. Finally, constraints (\ref{eq:c6}) and (\ref{eq:c7}) define the decision variables.

We also note that, in \cite{pancham-thesis}, the $d$-relaxed priority constraint was modeled by the following constraints:
\begin{align}
s_i &< s_j \qquad \forall i \in V_p, j \in V_q, q = p + d + 1 \label{eq:c8}
\end{align}

However, it can be observed that these constraints cannot model the $d$-relaxed rule defined in the last section. For example, consider a route with the sequence of priorities \{1, 2, 1, 3, 2, 4, 7, 3, 5, 6\} and $d = 2$. It is clear that this route satisfies the constraints (\ref{eq:c8}) but does not satisfy the $d$-relaxed constraint because from priority 4, we can only go to priority at most 3 + 2 = 5.

In the following, we slightly modify the formulation (F1) to propose a formulation called (F2). We replace the variables $s_i, i \in V$ in (F1) by the  new variables $u_i$ representing the order of node $v_i$ in the solution and add constraints (\ref{eq:c9})-(\ref{eq:c12}) to (F2).

\begin{align}
\textrm{Minimize} \, \qquad (\ref{eq:obj}) \nonumber \\
\textrm{subject to} \qquad \;\; u_1 &= 0 \label{eq:c13}\\
u_i + 1 - M(1-x_{ij}) &\leq u_j \qquad \forall \{i,j\} \in E; j \neq 1 \label{eq:c14}\\
u_i &< u_j \qquad \forall i \in V_p, j \in V_q, q > p + d \label{eq:c15}\\
\textrm{Constraints} (\ref{eq:c1}), (\ref{eq:c2}), &(\ref{eq:c6}), (\ref{eq:c9})-(\ref{eq:c12})\\
u_i &\in R  \qquad \forall i \in V \label{eq:c16}
\end{align}

Using $u$ variables instead of $s$ variables brings two advantages. First, it is easier to estimate the value of $M$ in (F2) than in (F1). Indeed, $M$ in $F2$ can be estimated simply by the number of nodes in the graph $n$ while in (F1), one can compute $M$ by constructing a feasible solution then take its objective value. Moreover, it is well-known that the big $M$ in MIPs can cause numerical issues and deteriorate the performance of exact methods based on MIP. With a more efficient estimation of $M$ in (F2), the performance of exact method based on (F2) should be better. In the computational section, the comparison of the performance of the two formulations will be conducted and analyzed.

The formulations above can be strengthen by adding the following constraints to create two models denoted by (F1') and (F2'), respectively:

\begin{align}
\sum_{i \in V_p} \sum_{j \in V_q} x_{ji} &= 0 \qquad \forall p, q; q > p + d \label{eq:c9}\\
\sum_{i \in V_p} x_{1i} &= 0 \qquad \forall p,q: p > 1 + d \label{eq:c10}\\
\sum_{i \in V_p} x_{i(n+1)} &= 0 \qquad \forall p,q: p < g - d \label{eq:c11}\\
\sum_{i \in V_p}\sum_{j \in V_q} x_{ij} &\leq 1 \qquad \forall p, q: q > p+d \label{eq:c12}
\end{align}

Constraints (\ref{eq:c9}) proposed in \cite{pancham-thesis} ensure that the vehicle can never travel directly from a priority $q$ node $j$ to a priority $p$ node $i$, when $q \geq p + 1 + d$. Additionally, we propose three new constraints (\ref{eq:c10})-(\ref{eq:c12}). Constraints (\ref{eq:c10}) and (\ref{eq:c11}) express the $d$-relaxed priority rule when we depart from and come back to the depot. Constraints (\ref{eq:c12}) say that vehicle will travel at most once from a priority $p$ node $i$ to a priority $q$ node $j$ when $q > p + d$. 

\section{A metaheuristic}\label{sec:solution}
In this section, we propose an approximate method to solve the problem. We use the framework called GILS-RVND that brings together the components of Greedy Randomized Adaptive Search Procedure (GRASP), Iterated Local Search (ILS) and Random Variable Neighborhood Descent (RVND). The approach developed by \cite{silva} was used to successfully solve the Minimum Latency Problem. The pseudo-code of the algorithm is presented in Algorithm \ref{algo:main}. Here, $f(s)$ denotes the objective value of a solution $s$. The algorithm starts with $I_{max}$ initial solutions which are generated using a greedy randomized procedure $Initialize()$. Each initial solution $s$ is then improved by means of a local search procedure $LocalSearch(s)$ combined with a perturbation mechanism $Pertubation(s)$ in the spirit of ILS. The local search procedure is designed according to the idea of RVND. It is important to note that the current best solution $s'$ is always selected to perform the perturbation at a given iteration (acceptance criterion). After $I_{ILS}$ iterations without improvements, the process is stopped and the search continues with another initial solution. 
\begin{algorithm}
\caption{GILS-RVND algorithm}\label{algo:main}
\begin{algorithmic}[1]
\Procedure{GILS-RVND$(I_{max}, I_{ILS})$}{}  
    \State $f^{*} \gets \infty $   
    \For{$i \gets 1,...,I_{max}$}
   		\State $s \gets Initialize()$
        \State $s' \gets s$
        \State $j \gets 0 $
        \While{$j < I_{ILS}$}
        	\State $s \gets LocalSearch(s)$           
            \If {$f(s) < f(s')$}
            	\State $s' \gets s$          
                \State $j \gets 0$
            \EndIf
        	\State $s \gets Perturbation(s')$
            \State $j \gets j + 1$
     	\EndWhile\label{ilsendwhile}  
        \If {$f(s') < f^{*}$}
    	\State $s^{*} \gets s'$
        \State $f^{*} \gets f(s')$
    \EndIf
    \EndFor   
	
    \State \textbf{return} $s^{*}$
\EndProcedure
\end{algorithmic}
\end{algorithm}

We now describe in more details the three main components of our algorithm: generation of an initial solution, local search, and perturbation:

\paragraph{\em Initial solutions:} The algorithm starts by generating initial solutions; each solution $s$ is constructed by adding the depot as the first visited location. At each iteration, a randomly selected node of $k$ nearest nodes with respect to the recently added node satisfying the $d$-relaxed priority rule, is inserted to the tour. To satisfy the $d$-relaxed priority rule, the candidate vertex must have a priority less than the summation of the minimum priority of unvisited vertices and $d$. The process is repeated until all the delivery locations are visited. The procedure is described in Algorithm \ref{algo:init}.
\begin{algorithm}
\caption{Procedure for creating initial solutions}\label{algo:init}
\begin{algorithmic}[1]
\Procedure{Initialize$(k)$}{} 	
	\State $s \cup   \{ 0 \} $ 
    \State Initialize Candidate List ($CL$) with all vertices
    \State $CL \gets CL - \{0\}$ 
    \State $r \gets 0$
    \While{$ CL \neq \emptyset$}
    	\State Sort $CL$ in ascending order according to their distance with respect to $r$;
    	\State Update Restricted Candidate List ($RCL$) considering only $k$ best candidates of $CL$ which satisfy $d$-relaxed rule.
    	\State Choose randomly $v$ $\in$ $RCL$ 
        \State $s \cup \{ v \}$    
        \State $r \gets v$
    	\State $CL \gets CL - \{r\}$
    \EndWhile
    \State \textbf{return} $s$
\EndProcedure
\end{algorithmic}
\end{algorithm}
 

\paragraph{\em Local search:} 

The local search is performed by a method based on RVND. Let $t$ be the number of neighborhood structures and $N = \{N^1, N^2, N^3,..., N^t \}$ be their corresponding set. Whenever a given neighborhood fails to improve the current best solution, RVND randomly selects another neighborhood to continue the search. The process stops if all the neighborhoods cannot improve the result. Preliminary tests revealed that this approach is capable of finding better solutions as compared to those that adopt a deterministic order. In our research, the set $N$ is composed of the following five neighborhood structures, whose associated solutions are explored in an exhaustive fashion with a best improvement strategy.

\begin{itemize}
\item \textbf{Relocate (1)} $- N^{(1)} - $ One node is relocated to another position of the tour .
\item \textbf{Relocate (2)} $- N^{(2)} - $ Two adjacent nodes are reallocated to another position of the tour.
\item \textbf{Swap (1-1)} $- N^{(3)} - $ Two nodes of the tour are interchanged.
\item \textbf{Swap (2-1)} $- N^{(4)} - $ Two adjacent nodes are interchanged with one node.
\item \textbf{Swap (2-2)} $- N^{(6)} - $ Two adjacent nodes of the tour are interchanged with another two adjacent nodes.
\end{itemize}

To check the feasibility of a local search move in $O(1)$, we use additional data structures $p^{min}_{ij}$ (and $p^{max}_{ij}$) presenting the minimum (and maximum) priority of nodes from $i^{th}$ location to $j^{th}$ location of the route $r$. In the following, we denote the $i^{th}$ node of route $r$ as $r[i]$. The conditions to check if a local search move satisfies the $d$-relaxed rule are as follows:
 
\textbf{Swap (1, 1)}:
We can swap 2 nodes $r[i]$ and $r[j]$ ($i < j$) of a route $r$ if
\begin{itemize}
\item There is no node with priority smaller than $p_{r[j]} - d$ appearing from  $i^{th}$ location to $(j-1)^{th}$ location: $ p^{min}_{i(j-1)} \geq p_{r[j]} - d$.
\item There is no node with priority greater than $p_{r[i]} + d$ appearing from $(i+1)^{th}$ location to $j^{th}$ location: $p_{r[i]} \geq p^{max}_{(i+1)j} - d$
\end{itemize}

\textbf{Swap (2, 1)}: In case of swapping an adjacent edge $(r[i], r[i+1])$ and a node $r[j]$ of a route $r$, two situations are considered: 

\begin{itemize}
\item If $j > i + 1$: There is no node with priority smaller than $p_{r[j]} - d$  appearing from  $i^{th}$ location to $(j-1)^{th}$ location: $ p^{min}_{i(j-1)} \geq p_{r[j]} - d$; and there is no node with priority greater than $p_{r[i]} + d$  or $p_{r[i+1]} + d$ appearing from $(i+2)^{th}$ location to $j^{th}$ location: $p^{max}_{(i+2)j} \leq \min(p_{r[i]},p_{r[i+1]}) + d$.
\item If $j < i$: There is no node with priority smaller than $p_{r[i]} - d$ or $p_{r[i+1]} - d$ appearing from  $j^{th}$ location to $(i-1)^{th}$ location: $ p^{min}_{j(i-1)} \geq \max(p_{r[i]},p_{r[i+1]}) - d$; and there is no node with priority greater than $p_{r[j]} + d$ appearing from $(j+1)^{th}$ location to $(i+1)^{th}$ location: $p^{max}_{(j+1)(i+1)} \leq p_{r[j]} + d$.
\end{itemize}

\textbf{Swap (2, 2)}: Similarly, we can swap an adjacent edge ($r[i], r[i+1]$) and an adjacent edge  $(r[j], r[j+1])$ ($i+1 < j$) of a route $r$ if 

\begin{itemize}
\item $p^{min}_{i(j-1)} \geq \max(p_{r[j]}, p_{r[j+1]}) - d$ 
\item $p^{max}_{(i+2)(j+1)} \leq \min(p_{r[i]}, p_{r[i+1]}) + d $ 
\end{itemize}

\textbf{Relocate(1)}: we can relocate a node $r[i]$ to right before $j^{th}$ location of a route $r$ if

\begin{itemize}
\item $p^{min}_{j(i-1)} \geq p_{r[i]} - d$ if $j < i$
\item $p^{max}_{(i+1)(j-1)} \leq p_{r[i]} + d $ if $i < j - 1$
\end{itemize}

\textbf{Relocate(2)}: we can relocate an adjacent edge $(r[i], r[i+1])$ to right before $j^{th}$ location of a route $r$ if

\begin{itemize}
\item $p^{min}_{j(i-1)} \geq \max(p_{r[i]},p_{r[i+1]}) - d$ if $j < i$
\item $p^{max}_{(i+2)(j-1)} \leq \min(p_{r[i]},p_{r[i+1]})$ + d if $j > i+2$ 
\end{itemize}




\paragraph{\em Perturbation:} The role of the perturbation is to lead the search to escape from local optima, when the local search fails to improve the current solution. The perturbation can be built by a random move in a neighborhood different from or larger than the ones used by the local search operators. The perturbation needs to be well tuned to be neither too weak nor too strong. In fact, if changes are too weak, the local search might bring us back to the previous solution. On the other hand, if changes are too strong, we could lose good attributes of the solution and perform random fruitless jumps in the solution space. Here, we use a simple and effective perturbation which includes $I_p$ feasible swap and relocate moves with selected probabilities of $p$ and (1-$p$) for each. By satisfying $d$-relaxed constraint, a feasible tour is generated. Experiments show that $p$ being set to 0.75 and $I_p$ to a random value from 4 to 6 (instances with $n < 60$), from 6 to 8 (instances with $60 \leq n \leq 100$), and 8 to 10 (instances with $n > 100$) lead to the best performance of the algorithm.


\section{Experimental results}\label{sec:report}
\subsection{Datasets and experimental settings}
\paragraph\
Due to the unavailability of datasets for the problem, we generate the benchmark instances based on TSP instances of TSPLIB \cite{TSPLIB95} by adding information on grouping procedures and $d$ value. The number of groups $g$ is set to one of three values \{1, 3, 5\}. Two grouping procedures are investigated: random and clustered. In the first type, a node is randomly distributed to a group. This could model applications in distribution commercial products in which the priority of a delivery node is determined by its storage level. The second type distributes the nodes closed to each other into a group and can imitate the situation in which natural disasters such as earthquake, tsunami affect areas along their propagation directions. To evaluate the impact of the $d$-relaxed priority rule, we tested instances with $d = 0, 1, 3$. The traveling cost and traveling time between two vertices are both set to their Euclidean distance. The instances are labeled X-Y-\textit{g}-\textit{d} where X is the name of original TSPLIB instances, Y represents the clustering procedure C (clustered) or R (random).

Our GILS-RVND approach is implemented in C++ language. All experiments are performed on a 64-bit machine with Intel(R) Core(TM) i7-6700 CPU at 3.40 GHz processor and 8.00 GB RAM. CPLEX 12.8.0 is used to solve the MIP formulations described above. 

\subsection{Results on small and medium-size instances}
We first evaluate the performance of our MIP formulations and GILS-RVND approach on a set of 42 and 52-node instances generated from swiss42 and berlin52 of TSPLIB. For each instance setting, three instances denoted with a letter a, b, c at the end of the label have randomly been generated. Four formulations are considered: F1, F1', F2 and F2'. We also run the metaheuristic with the fast setting $(I_{max}, I_{ILS}) = (5,500)$ to compare its solutions with best found upper bounds of CPLEX. The obtained results are presented in Table \ref{smallRandom} where  columns ``Sol" present the objective values of obtained solutions. Columns ``Gap" present the distance between lower and upper bounds returned by CPLEX and columns `Time" report the running time in seconds. 

As can be seen from Tables \ref{smallRandom} and \ref{table:smallClustered}, F2 can solve three more instance than F1, all in the class of random instances. The computational time of F2 on average is also less than that of F1. Moreover, adding four constraints (\ref{eq:c9}) -(\ref{eq:c12}) to the models F1 and F2 leads to the dominating performance of F2' over other F1', in both class of instances.  This shows the advantage of using $u$ variables instead of $s$ variables. But it is important to mention that on several instances mostly in the clustered class such as swiss42C-5-0-a and swiss42C-5-1-b, F1 is significantly faster than F2. The results also show the positive impact of four constraints (\ref{eq:c9}) -(\ref{eq:c12}) to the performance of the exact approaches when they help to solve more instances in shorter running time. An interesting observation is that instances with $d = \{1, 3\}$ are more difficult than the others. 

Another observation is that random instances seem to be harder to solve than clustered instances. Exact methods could solve to optimality more clustered instances than random instances. More precisely, the exact method based on the best formulation F2' can successfully solve 30/32 clustered instances but can solve only 22 random instances. 

Our metaheuristic provides promising results on the considering instances. It can retrieve all the optimal solutions in a shorter running time which is less than 15 seconds on all instances. Moreover, it improves the results of exact methods on the instance berlin52R-5-1-a.

\begin{table}[tbp]
\centering
\resizebox{\textwidth}{!}
{
 \begin{tabular}{|c|c|c|c|c|c|c|c|c|c|c|c|c|c|c|}
\hline
\multirow{2}{*}{} & \multicolumn{3}{c|}{\textbf{F1}} & \multicolumn{3}{c|}{\textbf{F'1}} & \multicolumn{3}{c|}{\textbf{F2}} & \multicolumn{3}{c|}{F'2} & \multicolumn{2}{l|}{\textbf{ILS 5x500}} \\ \cline{2-15} 
 & \textbf{Sol} & \textbf{Gap (\%)} & \textbf{Time} & \textbf{Sol}  & \textbf{Gap (\%)} & \textbf{Time} & \textbf{Sol}  & \textbf{Gap (\%)} & \textbf{Time} & \textbf{Sol} &  \textbf{Gap (\%)} & \textbf{Time} & \textbf{Sol} & \textbf{Time} \\ \hline
\textbf{swiss42R-1-0-a} & 1273 & 0 & 4.17 & 1273 & 0 & 1.59 & 1273 & 0 & 0.87 & 1273 & 0 & 0.87 & 1273 & 8.34 \\ \hline
\textbf{swiss42R-3-0-a} & 2256 & 0 & 3.42 & 2256 & 0 & 0.54 & 2256 & 0 & 0.8 & 2256 & 0 & 0.56 & 2256 & 4.35 \\ \hline
\textbf{swiss42R-3-0-b} & 2153 & 0 & 6.74 & 2153 & 0 & 1.29 & 2153 & 0 & 1.42 & 2153 & 0 & 1.35 & 2153 & 5.09 \\ \hline
\textbf{swiss42R-3-0-c} & 2080 & 0 & 10.34 & 2080 & 0 & 5.95 & 2080 & 0 & 2.34 & 2080 & 0 & 2.1 & 2080 & 5.08 \\ \hline
\textbf{swiss42R-3-1-a} & 1652 & 0 & 1168.96 & 1652 & 0 & 175.89 & 1652 & 0 & 1610.97 & 1652 & 0 & 103.45 & 1652 & 6.33 \\ \hline
\textbf{swiss42R-3-1-b} & 1607 & 1.85 & 3600.27 & 1607 & 0 & 689.18 & 1607 & 0 & 1378.93 & 1607 & 0 & 460.36 & 1607 & 5.74 \\ \hline
\textbf{swiss42R-3-1-c} & 1525 & 1.78 & 3600.23 & 1525 & 1.65 & 3602.05 & 1525 & 0.54 & 3600.14 & 1525 & 0 & 947.38 & 1525 & 6.14 \\ \hline
\textbf{swiss42R-5-0-a} & 2365 & 0 & 7.72 & 2365 & 0 & 1.34 & 2365 & 0 & 1.46 & 2365 & 0 & 1.33 & 2365 & 4.63 \\ \hline
\textbf{swiss42R-5-0-b} & 2567 & 0 & 5.37 & 2567 & 0 & 0.90 & 2567 & 0 & 1.47 & 2567 & 0 & 1.04 & 2567 & 4.36 \\ \hline
\textbf{swiss42R-5-0-c} & 2694 & 0 & 3.66 & 2694 & 0 & 1.01 & 2694 & 0 & 0.89 & 2694 & 0 & 0.8 & 2694 & 4.03 \\ \hline
\textbf{swiss42R-5-1-a} & 1812 & 0 & 892.2 & 1812 & 0 & 2585.26 & 1812 & 0 & 332.02 & 1812 & 0 & 219.82 & 1812 & 5.25 \\ \hline
\textbf{swiss42R-5-1-b} & 1905 & 3.81 & 3600.11 & 1964 & 9.05 & 3605.59 & 1905 & 2.59 & 3600.11 & 1905 & 4.34 & 3600.14 & 1905 & 5.14 \\ \hline
\textbf{swiss42R-5-1-c} & 1910 & 1.34 & 3600.28 & 1910 & 3.16 & 3605.14 & 1910 & 1.89 & 3600.11 & 1910 & 0 & 1845.06 & 1910 & 4.87 \\ \hline
\textbf{swiss42R-5-3-a} & 1509 & 8.17 & 3600.16 & 1474 & 0.84 & 3600.17 & 1498 & 6.53 & 3606.42 & 1509 & 4.55 & 3601.31 & 1474 & 6.32 \\ \hline
\textbf{swiss42R-5-3-b} & 1606 & 12.19 & 3602.84 & 1594 & 9.46 & 3601.79 & 1570 & 7.46 & 3604.44 & 1541 & 5.44 & 3600.07 & 1541 & 6.91 \\ \hline
\textbf{swiss42R-5-3-c} & 1513 & 3.21 & 3600.22 & 1510 & 0 & 1364.69 & 1510 & 0 & 1782.54 & 1510 & 0 & 1679.83 & 1510 & 6.3 \\ \hline
\textbf{berlin52R-1-0-a} & 7542 & 0 & 2.45 & 7542 & 0 & 1.96 & 7542 & 0 & 3.08 & 7542 & 0 & 3.15 & 7542 & 10.41 \\ \hline
\textbf{berlin52R-3-0-a} & 12765 & 0 & 8.33 & 12765 & 0 & 5.70 & 12765 & 0 & 10.35 & 12765 & 0 & 9.37 & 12765 & 7.2 \\ \hline
\textbf{berlin52R-3-0-b} & 12668 & 0 & 4.30 & 12668 & 0 & 3.78 & 12668 & 0 & 3.33 & 12668 & 0 & 4.06 & 12668 & 7.84 \\ \hline
\textbf{berlin52R-3-0-c} & 12483 & 0 & 31.4 & 12483 & 0 & 15.32 & 12483 & 0 & 9.02 & 12483 & 0 & 11.76 & 12483 & 9.81 \\ \hline
\textbf{berlin52R-3-1-a} & 9504 & 6.03 & 3607.41 & 9473 & 1.64 & 3603.82 & 9473 & 3.95 & 3602.73 & 9473 & 0 & 2486.71 & 9473 & 11.56 \\ \hline
\textbf{berlin52R-3-1-b} & 9442 & 4.00 & 3602.04 & 9419 & 2.58 & 3601.50 & 9419 & 3.49 & 3602.11 & 9442 & 2.39 & 3601.4 & 9419 & 10.14 \\ \hline
\textbf{berlin52R-3-1-c} & 9737 & 8.52 & 3608.44 & 9684 & 4.67 & 3601.37 & 9678 & 5.69 & 3603.63 & 9577 & 3.75 & 3601.67 & 9577 & 8.73 \\ \hline
\textbf{berlin52R-5-0-a} & 16414 & 0 & 4.37 & 16414 & 0 & 4.46 & 16414 & 0 & 2.5 & 16414 & 0 & 2.7 & 16414 & 8.93 \\ \hline
\textbf{berlin52R-5-0-b} & 13759 & 0 & 26.43 & 13759 & 0 & 49.97 & 13759 & 0 & 18.25 & 13759 & 0 & 24.55 & 13759 & 7.52 \\ \hline
\textbf{berlin52R-5-0-c} & 14131 & 0 & 20.92 & 14131 & 0 & 4.71 & 14132 & 0 & 5.62 & 14131 & 0 & 3.65 & 14131 & 7.56 \\ \hline
\textbf{berlin52R-5-1-a} & 11662 & 11.38 & 3609.71 & 12345 & 14.18 & 3602.87 & 11686 & 9.91 & 3608.77 & 12417 & 14.53 & 3605.1 & \textbf{11651} & 8.99 \\ \hline
\textbf{berlin52R-5-1-b} & 9957 & 8.93 & 3608.14 & 10158 & 8.26 & 3601.62 & 9982 & 6.95 & 3604.09 & 9982 & 6.6 & 3602.88 & 9957 & 7.31 \\ \hline
\textbf{berlin52R-5-1-c} & 10940 & 9.15 & 3608.94 & 11020 & 9.33 & 3600.11 & 10940 & 8.29 & 3601.56 & 11020 & 9.32 & 3603.29 & 10940 & 8.2 \\ \hline
\textbf{berlin52R-5-3-a} & 9803 & 15.89 & 3602.75 & 9065 & 7.39 & 3602.13 & 9065 & 7.73 & 3601.49 & 9085 & 6.97 & 3603.56 & 9065 & 9.38 \\ \hline
\textbf{berlin52R-5-3-b} & 8119 & 3.37 & 3606.91 & 8041 & 0.92 & 3600.23 & 8036 & 0 & 2994.48 & 8036 & 0 & 3387.42 & 8036 & 8.28 \\ \hline
\textbf{berlin52R-5-3-c} & 8224 & 3.17 & 3600.42 & 8224 & 2.37 & 3602.12 & 8274 & 3.5 & 3603.86 & 8224 & 2.02 & 3600.17 & 8224 & 8.48 \\ \hline

\end{tabular}
}
\caption{Performance of MIP models on random instances}
\label{smallRandom}
\end{table}

\begin{table}[tbp]
\centering
\resizebox{\textwidth}{!}
{
 \begin{tabular}{|c|c|c|c|c|c|c|c|c|c|c|c|c|c|c|}
\hline
\multirow{2}{*}{} & \multicolumn{3}{c|}{\textbf{F1}} & \multicolumn{3}{c|}{\textbf{F'1}} & \multicolumn{3}{c|}{\textbf{F2}} & \multicolumn{3}{c|}{F'2} & \multicolumn{2}{l|}{\textbf{ILS 5x500}} \\ \cline{2-15} 
 & \textbf{Sol} & \textbf{Gap (\%)} & \textbf{Time} & \textbf{Sol}  & \textbf{Gap (\%)} & \textbf{Time} & \textbf{Sol}  & \textbf{Gap (\%)} & \textbf{Time} & \textbf{Sol} &  \textbf{Gap (\%)} & \textbf{Time} & \textbf{Sol} & \textbf{Time} \\ \hline
\textbf{swiss42C-1-0} & 1273 & 0 & 1.05 & 1273 & 0 & 1.1 & 1273 & 0 & 0.85 & 1273 & 0 & 0.82 & 1273 & 7.98 \\ \hline
\textbf{swiss42C-3-0-a} & 1347 & 0 & 3.07 & 1347 & 0 & 2.56 & 1347 & 0 & 3.21 & 1347 & 0 & 1.94 & 1347 & 4.31 \\ \hline
\textbf{swiss42C-3-0-b} & 1505 & 0 & 6.45 & 1505 & 0 & 9.88 & 1505 & 0 & 9.92 & 1505 & 0 & 9.22 & 1505 & 4.50 \\ \hline
\textbf{swiss42C-3-0-c} & 1467 & 0 & 4.03 & 1467 & 0 & 1.8 & 1467 & 0 & 2.15 & 1467 & 0 & 1.67 & 1467 & 4.16 \\ \hline
\textbf{swiss42C-3-1-a} & 1301 & 0 & 7.78 & 1301 & 0 & 3.31 & 1301 & 0 & 5.14 & 1301 & 0 & 2.79 & 1301 & 4.75 \\ \hline
\textbf{swiss42C-3-1-b} & 1344 & 0 & 9.60 & 1344 & 0 & 4.43 & 1344 & 0 & 6.41 & 1344 & 0 & 3.23 & 1344 & 4.69 \\ \hline
\textbf{swiss42C-3-1-c} & 1357 & 0 & 28.69 & 1357 & 0 & 2.79 & 1357 & 0 & 13.05 & 1357 & 0 & 2.31 & 1357 & 4.25 \\ \hline
\textbf{swiss42C-5-0-a} & 1561 & 0 & 569.88 & 1561 & 0 & 261.76 & 1561 & 0 & 2330.89 & 1561 & 0 & 170.83 & 1561 & 3.47 \\ \hline
\textbf{swiss42C-5-0-b} & 1540 & 0 & 3.14 & 1540 & 0 & 2.3 & 1540 & 0 & 4.99 & 1540 & 0 & 2.14 & 1540 & 3.48 \\ \hline
\textbf{swiss42C-5-0-c} & 1532 & 0 & 46.51 & 1532 & 0 & 17.82 & 1532 & 0 & 43.48 & 1532 & 0 & 13.28 & 1532 & 3.39 \\ \hline
\textbf{swiss42C-5-1-a} & 1434 & 0 & 17.69 & 1434 & 0 & 23.02 & 1434 & 0 & 13.40 & 1434 & 0 & 14.53 & 1434 & 5.19 \\ \hline
\textbf{swiss42C-5-1-b} & 1469 & 0 & 1142.43 & 1469 & 0 & 48.62 & 1469 & 0 & 1352.94 & 1469 & 0 & 32.79 & 1469 & 4.85 \\ \hline
\textbf{swiss42C-5-1-c} & 1334 & 0 & 1.73 & 1334 & 0 & 1.66 & 1334 & 0 & 1.93 & 1334 & 0 & 2.07 & 1334 & 4.57 \\ \hline
\textbf{swiss42C-5-3-a} & 1273 & 0 & 0.95 & 1273 & 0 & 0.88 & 1273 & 0 & 0.84 & 1273 & 0 & 0.93 & 1273 & 4.84 \\ \hline
\textbf{swiss42C-5-3-b} & 1273 & 0 & 1.36 & 1273 & 0 & 0.82 & 1273 & 0 & 1.05 & 1273 & 0 & 1.53 & 1273 & 5.14 \\ \hline
\textbf{swiss42C-5-3-c} & 1301 & 0 & 6.02 & 1301 & 0 & 4.05 & 1301 & 0 & 5.85 & 1301 & 0 & 5.88 & 1301 & 5.91 \\ \hline
\textbf{berlin52C-1-0} & 7542 & 0 & 4.29 & 7542 & 0 & 3.17 & 7542 & 0 & 3.24 & 7542 & 0 & 2.45 & 7542 & 8.58 \\ \hline
\textbf{berlin52C-3-0-a} & 8144 & 0 & 3.26 & 8144 & 0 & 2.31 & 8144 & 0 & 3.14 & 8144 & 0 & 1.82 & 8144 & 9.68 \\ \hline
\textbf{berlin52C-3-0-b} & 8016 & 0 & 10.88 & 8016 & 0 & 2.85 & 8016 & 0 & 10.85 & 8016 & 0 & 2.95 & 8016 & 6.70 \\ \hline
\textbf{berlin52C-3-0-c} & 9085 & 0 & 502.10 & 9085 & 0 & 339.78 & 9085 & 0 & 245.93 & 9085 & 0 & 270.75 & 9085 & 7.44 \\ \hline
\textbf{berlin52C-3-1-a} & 7952 & 0 & 49.63 & 7952 & 0 & 5.99 & 7952 & 0 & 32.27 & 7952 & 0 & 3.72 & 7952 & 9.08 \\ \hline
\textbf{berlin52C-3-1-b} & 7596 & 0 & 6.13 & 7596 & 0 & 2.29 & 7596 & 0 & 3.66 & 7596 & 0 & 3.31 & 7596 & 7.27 \\ \hline
\textbf{berlin52C-3-1-c} & 7984 & 0 & 2.74 & 7984 & 0 & 3.01 & 7984 & 0 & 2.98 & 7984 & 0 & 1.50 & 7984 & 9.20 \\ \hline
\textbf{berlin52C-5-0-a} & 9430 & 0 & 85.95 & 9430 & 0 & 34.47 & 9430 & 0 & 50.94 & 9430 & 0 & 68.18 & 9430 & 7.53 \\ \hline
\textbf{berlin52C-5-0-b} & 8669 & 0 & 12.33 & 8669 & 0 & 9.65 & 8669 & 0 & 8.28 & 8669 & 0 & 10.37 & 8669 & 5.86 \\ \hline
\textbf{berlin52C-5-0-c} & 9651 & 6.03 & 3601.55 & 9651 & 6.07 & 3,601.89 & 9651 & 6.08 & 3601.59 & 9651 & 5.97 & 3602.66 & 9651 & 6.69 \\ \hline
\textbf{berlin52C-5-1-a} & 8811 & 0 & 5.92 & 8811 & 0 & 3.25 & 8811 & 0 & 10.25 & 8811 & 0 & 3.53 & 8811 & 8.57 \\ \hline
\textbf{berlin52C-5-1-b} & 7948 & 0 & 13.07 & 7948 & 0 & 2.69 & 7948 & 0 & 4.59 & 7948 & 0 & 3.00 & 7948 & 7.38 \\ \hline
\textbf{berlin52C-5-1-c} & 8509 & 0 & 11.98 & 8509 & 0 & 18.61 & 8509 & 0 & 10.29 & 8509 & 0 & 11.95 & 8509 & 7.94 \\ \hline
\textbf{berlin52C-5-3-a} & 7907 & 1.24 & 3600.16 & 7907 & 1.04 & 3,600.16 & 7907 & 1.19 & 3600.14 & 7907 & 1.08 & 3600.18 & 7907 & 11.99 \\ \hline
\textbf{berlin52C-5-3-b} & 7614 & 0 & 8.71 & 7614 & 0 & 1.91 & 7614 & 0 & 5.04 & 7614 & 0 & 1.81 & 7614 & 8.43 \\ \hline
\textbf{berlin52C-5-3-c} & 7631 & 0 & 10.82 & 7631 & 0 & 3.57 & 7631 & 0 & 3.03 & 7631 & 0 & 2.13 & 7631 & 12.20 \\ \hline
\end{tabular}
}
\caption{Performance of MIP models on clustered instances}
\label{table:smallClustered}
\end{table}

\subsection{Results of GILS-RVND on larger instances}
 A set of CTSP-$d$ instances with 100 and 200 nodes based on kroA, kroB, kroC, kroD in TSPLIB was generated to analyze the efficiency of the metaheuristic on larger instances. The metaheuristic is run 10 times on each instance with setting $(I_{max}, I_{ILS}) = (5,1000)$. The notations used in Tables \ref{largeRandom} and \ref{largeClustered} are explained as follows:\
\begin{itemize}
\item Instance: the name of instance
\item AvgSol: the objective value of solutions on average over 10 runs.
\item AvgTime: averaged running time over 10 runs
\item AvgGap: the gap on average of solutions to the best found solution over 10 runs.
\item BKS: the best known solution found over 10 runs 
\item GapTSP: the gap of the best found solution to the optimal TSP solution.
\end{itemize}

\begin{table}[tbp]
\centering
\resizebox{\textwidth}{!}
{\begin{tabular}{|c|c|c|c|c|c|c|c|c|c|c|c|}
\hline
\textbf{Instance} & \textbf{AvgSol} & \textbf{AvgTime} & \textbf{AvgGap} & \textbf{BKS}& \textbf{GapTSP} & \textbf{Instance} & \textbf{AvgSol} & \textbf{AvgTime} & \textbf{AvgGap} & \textbf{BKS} & \textbf{GapTSP} \\ \hline
\textbf{kroA100-R-1-0} & 21365.5 & 95.23 & 0.39 & 21282 & 0.00 & \textbf{kroA100-C-1-0} & 21338.6 & 84.88 & 0.26 & 21282 & 0.00 \\ \hline
\textbf{kroA100-R-3-0} & 38877.5 & 73.43 & 0.16 & 38814 & 45.17 & \textbf{kroA100-C-3-0} & 24049 & 67.93 & 0.00 & 24049 & 11.51 \\ \hline
\textbf{kroA100-R-3-1} & 29578.2 & 92.64 & 1.05 & 29264 & 27.28 & \textbf{kroA100-C-3-1} & 23416.7 & 80.07 & 1.48 & 23069 & 7.75 \\ \hline
\textbf{kroA100-R-5-0} & 50411 & 63.04 & 0.43 & 50192 & 57.60 & \textbf{kroA100-C-5-0} & 24745 & 58.51 & 0.00 & 24745 & 13.99 \\ \hline
\textbf{kroA100-R-5-1} & 36328.4 & 72.97 & 1.32 & 35847 & 40.63 & \textbf{kroA100-C-5-1} & 22591.8 & 84.44 & 0.01 & 22589 & 5.79 \\ \hline
\textbf{kroA100-R-5-3} & 25594.9 & 90.97 & 0.87 & 25370 & 16.11 & \textbf{kroA100-C-5-3} & 21443 & 82.50 & 0.00 & 21443 & 0.75 \\ \hline
\textbf{kroB100-R-1-0} & 22189.1 & 83.55 & 0.22 & 22141 & 0.00 & \textbf{kroB100-C-1-0} & 22235 & 83.07 & 0.25 & 22179 & 0.17 \\ \hline
\textbf{kroB100-R-3-0} & 37770.2 & 68.96 & 0.17 & 37706 & 41.28 & \textbf{kroB100-C-3-0} & 24971.3 & 63.46 & 0.33 & 24887 & 11.03 \\ \hline
\textbf{kroB100-R-3-1} & 28652.6 & 85.82 & 0.5 & 28509 & 22.34 & \textbf{kroB100-C-3-1} & 22155.1 & 77.19 & 0.06 & 22141 & 0.00 \\ \hline
\textbf{kroB100-R-5-0} & 50863.4 & 57.83 & 0.16 & 50781 & 56.40 & \textbf{kroB100-C-5-0} & 24794 & 57.91 & 0.00 & 24794 & 10.70 \\ \hline
\textbf{kroB100-R-5-1} & 35589.3 & 81.01 & 1.06 & 35209 & 37.12 & \textbf{kroB100-C-5-1} & 23173.1 & 73.43 & 0.06 & 23159 & 4.40 \\ \hline
\textbf{kroB100-R-5-3} & 26205 & 85.42 & 0.51 & 26069 & 15.07 & \textbf{kroB100-C-5-3} & 22180 & 79.72 & 0.18 & 22141 & 0.00 \\ \hline
\textbf{kroC100-R-1-0} & 20826.1 & 94.55 & 0.37 & 20749 & 0.00 & \textbf{kroC100-C-1-0} & 20786.9 & 94.88 & 0.18 & 20749 & 0.00 \\ \hline
\textbf{kroC100-R-3-0} & 38083 & 86.17 & 0.34 & 37953 & 45.33 & \textbf{kroC100-C-3-0} & 21440.6 & 57.97 & 0.47 & 21340 & 2.77 \\ \hline
\textbf{kroC100-R-3-1} & 28304.5 & 86.83 & 0.61 & 28130 & 26.24 & \textbf{kroC100-C-3-1} & 20910 & 92.46 & 0.00 & 20910 & 0.77 \\ \hline
\textbf{kroC100-R-5-0} & 50099.9 & 69.09 & 0.03 & 50085 & 58.57 & \textbf{kroC100-C-5-0} & 24040 & 58.26 & 0.00 & 24040 & 13.69 \\ \hline
\textbf{kroC100-R-5-1} & 34365.1 & 73.73 & 2.22 & 33594 & 38.24 & \textbf{kroC100-C-5-1} & 22827 & 71.76 & 0.00 & 22827 & 9.10 \\ \hline
\textbf{kroC100-R-5-3} & 25587.6 & 94.83 & 0.5 & 25458 & 18.50 & \textbf{kroC100-C-5-3} & 21344.9 & 91.31 & 0.31 & 21278 & 2.49 \\ \hline
\textbf{kroD100-R-1-0} & 21336.6 & 101.54 & 0.2 & 21294 & 0.00 & \textbf{kroD100-C-1-0} & 21298.5 & 93.39 & 0.02 & 21294 & 0.00 \\ \hline
\textbf{kroD100-R-3-0} & 38290.3 & 72.38 & 0.47 & 38110 & 44.12 & \textbf{kroD100-C-3-0} & 23833.3 & 65.25 & 0.10 & 23809 & 10.56 \\ \hline
\textbf{kroD100-R-3-1} & 27886.3 & 89.45 & 0.54 & 27734 & 23.22 & \textbf{kroD100-C-3-1} & 22268.9 & 87.50 & 1.44 & 21944 & 2.96 \\ \hline
\textbf{kroD100-R-5-0} & 49222.8 & 67.1 & 0.25 & 49100 & 56.63 & \textbf{kroD100-C-5-0} & 28234.9 & 56.30 & 0.02 & 28228 & 24.56 \\ \hline
\textbf{kroD100-R-5-1} & 34414.9 & 71.56 & 0.48 & 34246 & 37.82 & \textbf{kroD100-C-5-1} & 25250.8 & 72.72 & 0.59 & 25102 & 15.17 \\ \hline
\textbf{kroD100-R-5-3} & 25733.4 & 86.38 & 0.42 & 25624 & 16.90 & \textbf{kroD100-C-5-3} & 21747 & 94.90 & 0.01 & 21744 & 2.07 \\ \hline
\textbf{kroE100-R-1-0} & 22129.1 & 100.05 & 0.28 & 22068 & 0.00 & \textbf{kroE100-C-1-0} & 22146.9 & 89.83 & 0.35 & 22068 & 0.00 \\ \hline
\textbf{kroE100-R-3-0} & 37996 & 67.78 & 0.16 & 37935 & 41.83 & \textbf{kroE100-C-3-0} & 24405.4 & 66.54 & 0.09 & 24383 & 9.49 \\ \hline
\textbf{kroE100-R-3-1} & 29489.2 & 90.94 & 0.44 & 29359 & 24.83 & \textbf{kroE100-C-3-1} & 22125 & 76.26 & 0.02 & 22121 & 0.24 \\ \hline
\textbf{kroE100-R-5-0} & 54285.2 & 66.37 & 0.16 & 54197 & 59.28 & \textbf{kroE100-C-5-0} & 26443.3 & 57.50 & 0.01 & 26440 & 16.54 \\ \hline
\textbf{kroE100-R-5-1} & 38700.8 & 76.58 & 0.88 & 38359 & 42.47 & \textbf{kroE100-C-5-1} & 23658.1 & 62.91 & 0.20 & 23611 & 6.54 \\ \hline
\textbf{kroE100-R-5-3} & 27316.3 & 109.86 & 0.22 & 27256 & 19.03 & \textbf{kroE100-C-5-3} & 22560.6 & 76.54 & 0.47 & 22455 & 1.72 \\ \hline

\end{tabular}}
\caption{Results of GILS-RVND for instances with 100 deliveries}
\label{largeRandom}
\end{table}

\begin{table}[tbp]
\centering
\resizebox{\textwidth}{!}
{\begin{tabular}{|c|c|c|c|c|c|c|c|c|c|c|c|}
\hline
\textbf{Instance} & \textbf{AvgSol} & \textbf{AvgTime} & \textbf{AvgGap} & \textbf{BKS} & \textbf{GapTSP} & \textbf{Instance} & \textbf{AvgSol} & \textbf{AvgTime} & \textbf{AvgGap} & \textbf{BKS} & \textbf{GapTSP}  \\ \hline
\textbf{kroA200-R-1-0} & 30176.5 & 656.14 & 1.07 & 29853 & 1.62 & \textbf{kroA200-C-1-0} & 30042.8 & 641.03 & 1.01 & 29737 & 1.24 \\ \hline
\textbf{kroA200-R-3-0} & 52237.3 & 473.42 & 0.95 & 51741 & 43.24 & \textbf{kroA200-C-3-0} & 30090 & 607.22 & 0.58 & 29913 & 1.82 \\ \hline
\textbf{kroA200-R-3-1} & 38471.5 & 574.14 & 0.68 & 38208 & 23.14 & \textbf{kroA200-C-3-1} & 29862.4 & 612.36 & 1.42 & 29435 & 0.23 \\ \hline
\textbf{kroA200-R-5-0} & 67757.6 & 435.56 & 0.97 & 67096 & 56.23 & \textbf{kroA200-C-5-0} & 32273.3 & 404.43 & 0.15 & 32224 & 8.86 \\ \hline
\textbf{kroA200-R-5-1} & 48790.1 & 537.11 & 1.57 & 48020 & 38.84 & \textbf{kroA200-C-5-1} & 31221.3 & 467.58 & 0.49 & 31069 & 5.47 \\ \hline
\textbf{kroA200-R-5-3} & 34630.4 & 495.08 & 1.24 & 34195 & 14.12 & \textbf{kroA200-C-5-3} & 30964 & 565.14 & 0.89 & 30686 & 4.30 \\ \hline
\textbf{kroB200-R-1-0} & 30149.7 & 697.39 & 0.99 & 29849 & 1.38 & \textbf{kroB200-C-1-0} & 30064.7 & 628.30 & 0.91 & 29790 & 1.18 \\ \hline
\textbf{kroB200-R-3-0} & 54131.3 & 525.52 & 0.72 & 53739 & 45.22 & \textbf{kroB200-C-3-0} & 31276.1 & 565.49 & 0.91 & 30989 & 5.01 \\ \hline
\textbf{kroB200-R-3-1} & 39190.5 & 606.27 & 0.63 & 38943 & 24.41 & \textbf{kroB200-C-3-1} & 30830.7 & 539.31 & 1.21 & 30457 & 3.35 \\ \hline
\textbf{kroB200-R-5-0} & 73011.3 & 469.76 & 0.31 & 72786 & 59.56 & \textbf{kroB200-C-5-0} & 37973.2 & 401.54 & 0.17 & 37909 & 22.35 \\ \hline
\textbf{kroB200-R-5-1} & 50821.8 & 525.85 & 1.1 & 50260 & 41.43 & \textbf{kroB200-C-5-1} & 33438.3 & 540.77 & 0.48 & 33276 & 11.54 \\ \hline
\textbf{kroB200-R-5-3} & 37145.9 & 652.27 & 0.59 & 36926 & 20.28 & \textbf{kroB200-C-5-3} & 30492 & 565.99 & 0.73 & 30270 & 2.75 \\ \hline

\end{tabular}}
\caption{Results of GILS-RVND for instances with 200 deliveries}
\label{largeClustered}
\end{table}


The results show the stability of our metaheuristic as the averaged gap AvgGap never exceeds 2\%. Although there is no reference algorithm in the literature, we can still analyse the performance of our metaheuristic based on the results of instances with one group and $d$ being set to 0; or in other words on classical TSP instances whose optimal solutions are known. Our algorithm can find all TSP optimal solutions but one (instance kroB100-C-1-0) on instances with 100 deliveries. However, it cannot find any TSP optimal solution on instances with 200 delivery vertices. Its running time is acceptable, less than 2 minutes on 100-node instances and 
approximately 10 minutes on 200-node instances.

The objective value of solutions on the clustered instances tends to be smaller than on the random instances. This can be easily explained by the way the instances are generated. On the clustered instances, the delivery nodes in a group are closed to each other, and the $d$-relaxed priority constraint tends to enforce solution tours closed to TSP tours.

As can be seen in the result tables, the increase of $d$ leads to significant reduction in the travel cost. The reduction on the random instances is larger than that on the clustered instances.

\section{CONCLUSION AND FUTURE WORKS}\label{sec:conclusion}
\paragraph\
In this paper, we study the clustered traveling salesman problem with $d$-relaxed priority rule (CTSP-$d$). We improve the MIP model proposed in \cite{pancham-thesis} and propose the first approximate approach that uses a GILS-RVND framework introduced in \cite{silva} to solve this problem. The results show that our best formulation can solve instances up to 52 nodes. There are two indications of a good metaheuristic shown by the GILS-RVND. It can find all the optimal solutions and provide stable solutions in a reasonable running time on large instances.

Future research directions include improving the exact method by adding efficient valid inequalities in a branch-and-cut framework to solve larger instances. The metaheuristic based on larger neighborhoods such as Large Neighborhood Search (LNS) could be a good candidate for an improved approximate method. The capacitated version of the problem is also an interesting topic for future researches.

\paragraph{Acknowledgment:}
This research is funded by the Fonds de Recherche Québec Nature et Technologies, (FRQNT), Canada, under grant number 2015-PR-181381 and the Vietnam National Foundation for Science and Technology Development (NAFOSTED) under grant number 102.99-2016.21.

\pagebreak
\def\bibfont{\tiny}


\begin{thebibliography}{}
 
\bibitem{vr} 
P. Toth and D. Vigo.
\textit{ Vehicle Routing: Problems, Methods and Applications. MOS-SIAM}. 
Series in Optimization, Second edition, 2014

\bibitem{vrp_adv} 
B. L. Golden, S. Raghavan, and E. A. Wasil.
\textit{The vehicle routing problem: latest advances and new challenges}. 
volume 43. Springer, 2008.
 





\bibitem{tsp}
C. Rego, D. Gamboa, F. Glover, C. Osterman.
\textit{Traveling salesman problem heuristics: leading methods, implementations and latest advances}. 
European Journal of Operational Research, 211 (3): 427–441, 2011.







\bibitem{pancham-thesis}
K.V. Panchamgam. \textit{Essays in Retail Operations and Humanitarian Logistics}. Ph.D thesis, University of Maryland College Park, 2011. 

\bibitem{pancham-paper}
K. Panchamgam, Y. Xiong, B. Golden, B. Dussault, and E. Wasil. \textit{The Hierarchical Traveling Salesman Problem}. Optimization Letters, Volume 7, Issue 7: 1517-1524, October 2013.

\bibitem{vrpbook}
P. Toth, D. Vigo. \textit{Vehicle Routing: Problems, methods and applications}. MOS-SIAM Series on Optimization.

\bibitem{2003}
H. R. Lourenço, O. C. Martin, T. Stützle.
\textit{Iterated local search}. Handbook of Metaheuristics. Kluwer Academic Publishers, International Series in Operations Research \& Management Science. 57: 321-353, 2003.
\bibitem{bau} E.B. Baum. 
\textit{Iterated descent: a better algorithm for local search in combinatorial optimization}. Technical Report, Caltech, Pasadena, CA, 1986.
\bibitem{shop}
T. Stützle. 
\textit{Applying iterated local search to the permutation flow shop problem}. Technical Report AIDA–98–04, FG Intellektik, TU Darmstadt, Darmstadt, Germany, August 1998.

\bibitem{39}
H. Hashimoto, M. Yagiura, and T. Ibaraki.
\textit{An iterated local search algorithm for the time-dependent vehicle routing problem with time windows}.  Discrete Optimization, 5(2):434–456, 2008.
\bibitem{44}
T. Ibaraki, S. Imahori, K. Nonobe, K. Sobue, T. Uno, and M. Yagiura.
\textit{An iterated local search algorithm for the vehicle routing problem with convex time penalty functions}. Discrete Applied Mathematics, 156(11):2050–2069, 2008.
\bibitem{51}
B.W. Kernighan and S. Lin. Bell.
\textit{An efficient heuristic procedure for partitioning graphs}. Systems Technology Journal, 49:213–219, 1970.

\bibitem{54}
B. Laurent and J.-K. Hao.
\textit{Iterated local search for the multiple depot vehicle scheduling problem}. Computers \& Industrial Engineering,  57(1):277–286, 2009.


\bibitem{87}
L. Tang and X. Wang. 
\textit{Iterated local search algorithm based on a very large-scale neighborhood for prize-collecting vehicle routing problem}. The International Journal of Advanced Manufacturing Technology, 29(11–12):1246–1258, 2006.

\bibitem{TSPLIB95}
G. Reinelt.
\textit{TSPLIB - a traveling salesman problem library}.  ORSA Journal on Computing, 3:376–384, 1991.

\bibitem{silva}
M.S. Marcos, S. Anand, V. Thibaut Vidal, S.O. Luiz,
\textit{A simple and effective metaheuristic for the Minimum Latency Problem}. 221(3): 513-520, 2012.

\bibitem{ahmed}
Z.H. Ahmed. \textit{The ordered clustered traveling salesman problem: a hybrid genetic algorithm}. The Scientific World Journal, 2014.


\bibitem{chisman}
J.A. Chisman. \textit{The clustered traveling salesman problem}. Computers \& Operations Research. 2(2): 115-119, 1975.

\bibitem{mestria2018}
M. Mestria. \textit{New hybrid heuristic algorithm for the clustered traveling salesman problem}. Computers \& Industrial Engineering. 116: 1-12, 2018.

\bibitem{mestria2013}
M. Mestria, L.S. Ochi, S. de Lima Martins. \textit{Grasp with path relinking for the symmetric euclidean clustered traveling salesman problem}.  Computers \& Operations Research, 40 (12): 3218-3229, 2013.

\bibitem{zhang}
T. Zhang, L. Ke, J. Li, J. Li, J. Huang, Z. Li. \textit{Metaheuristics for the tabu clustered traveling salesman problem.} Computers \& Operations Research. 89: 1-12, 2013.

\bibitem{potvin1998}
J.Y. Potvin, F. Guertin. \textit{A Genetic Algorithm for the Clustered Traveling Salesman Problem with a Prespecified Order on the Clusters}. In: Woodruff D.L. (eds) Advances in Computational and Stochastic Optimization, Logic Programming, and Heuristic Search. Operations Research/Computer Science Interfaces Series, vol 9. Springer, Boston, MA, 1998.


\bibitem{potvin1996}
J-Y. Potvin, F. Guertin. The clustered traveling salesman problem: A genetic approach. Meta-Heuristics, 619-631, 1996.

\bibitem{laporte}
G. Laporte, J-Y. Potvin, F. Quilleret. \textit{A tabu search heuristic using genetic diversification for the clustered traveling salesman problem}. Journal of Heuristics, 2(3): 187–200, 1997.

\end{thebibliography}
\end{document}